%% file: main.tex
\documentclass[10pt,twocolumn,letterpaper]{article}

\usepackage{cvpr}

\usepackage{times}
\usepackage{latexsym}
\usepackage[T1]{fontenc}
\usepackage[utf8]{inputenc}
\usepackage{microtype}
\usepackage{graphicx}
\usepackage{amsmath}
\usepackage{amssymb}
\usepackage{booktabs}
\usepackage{multirow}
\usepackage{xcolor}
\usepackage{array}
\usepackage{enumitem}
\usepackage{float}

\usepackage[pagebackref,breaklinks,colorlinks,allcolors=blue]{hyperref}

\usepackage[capitalize]{cleveref}

\def\paperID{****}
\def\confName{CVPR}
\def\confYear{2026}

\providecommand{\keywords}[1]{}

\title{AffectFuse: Cross-Task Feature Fusion with Temporal Modeling for Multi-Task Affective Behavior Analysis}

\renewcommand*{\thefootnote}{\fnsymbol{footnote}}

\author{%
Dipit Saha\thanks{Equal contribution.} \quad Mohammad Raihan Rashid\footnotemark[1] \quad Shah Mohammad Abdul Mannan\footnotemark[1]\\
Ahnaf Tahmid\footnotemark[1] \quad Md.~Mehedi Hasan\footnotemark[1] \quad M.~Saifur Rahman\thanks{Corresponding author.}\\
Bangladesh University of Engineering and Technology\\
{\small\texttt{\{2105050,2105046,2105056,2105041,2105052\}@ugrad.cse.buet.ac.bd}}\\
{\small\texttt{mrahman@cse.buet.ac.bd}}\\
}

\begin{document}
\maketitle
\renewcommand*{\thefootnote}{\arabic{footnote}}
\setcounter{footnote}{0}

\input{sections/00-abstract}
\input{sections/01-introduction}
\input{sections/02-related-work}
\input{sections/03-method}
\input{sections/04-experimental-setup}
\input{sections/06-conclusion}

{
\small
\bibliographystyle{ieeenat_fullname}
\bibliography{refs}
}

\end{document}

%% file: sections/00-abstract.tex
\begin{abstract}
Affective behavior recognition in the wild requires joint prediction of
continuous valence-arousal, categorical facial expression, and multi-label
action units from unconstrained face images. We present our system for the
Multi-Task Learning (MTL) track of the 11th Affective Behavior Analysis
in-the-wild (ABAW) competition on \mbox{s-Aff-Wild2}, the static selected-frame
version of Aff-Wild2. The method focuses on post-encoder adaptation: frozen
AffectNet-supervised backbones provide multi-resolution features, while
task-specific temporal heads and cross-task fusion modules select the useful
signals for each target. For action-unit recognition, we adapt MAE-Face with
Low-Rank Adaptation (LoRA) and use DISFA
through per-unit expert routing rather than direct sequential transfer. Ablations
over backbone, temporal, fusion, and AU-adaptation choices define the final
configuration. The final system obtains $\mathcal{P}=1.7302$
on the official validation split, showing that post-encoder adaptation and
task-wise modeling choices provide a strong MTL pipeline without training a
new large-scale face foundation model. Code is available
\href{https://github.com/hackfleet-dev/affect-fuse-11th-abaw-mtl-challenge}{here}.
\keywords{Affective behavior analysis \and Multi-task learning \and Action unit
detection \and Valence--arousal \and Cross-task fusion \and Aff-Wild2.}
\end{abstract}

%% file: sections/01-introduction.tex
\section{Introduction}
\label{sec:intro}
The Affective Behavior Analysis in-the-wild (ABAW) competition
series~\cite{kollias2022abaw,kollias2023abaw2,kollias20246th,kollias20247th,kollias2025advancements,kollias2026affect}
studies emotion recognition from faces recorded in unconstrained conditions. Its
Multi-Task Learning (MTL) track ties three related problems together on the
static selected-frame \mbox{s-Aff-Wild2} subset of
Aff-Wild2~\cite{zafeiriou2017aff,kollias2019deep,kollias2019expression}: on
roughly 221K frames, one model must regress continuous \emph{valence} and
\emph{arousal}~\cite{russell1980circumplex}, classify one of eight \emph{expressions} (the six basic emotions,
\emph{Neutral}, and a broad \emph{Other} state), and detect twelve
\emph{action units} (AUs)~\cite{ekman1978facial}. We denote these tasks as
valence--arousal (VA), expression recognition (EXPR), and AU detection. A single number $\mathcal{P}$ (\cref{eq:candidate-selection}) sums the
mean valence/arousal concordance correlation coefficient, the macro-$F_1$ over
the eight expressions, and the mean $F_1$ over the twelve action units. The three
tasks reward different structure: expression is a local, discrete identity;
valence and arousal drift slowly; and action-unit detection is a sparse
multi-label problem dominated by rare units.

Our system focuses on the stages after feature extraction. We use frozen,
publicly available AffectNet-supervised face encoders, mainly EfficientNet-B2
and DDAMFN, then build task-specific candidate heads on top of their features.
The final configuration is selected through ablation studies over backbone,
fusion, temporal, and AU-adaptation choices. Three observations shape this design. First, these frozen supervised features are
already strong for expression and valence/arousal, so these tasks benefit more
from temporal modeling and cross-task fusion than from retraining the encoder.
Second, action units are the task where frozen features stall, so we adapt
MAE-Face with LoRA under a collapse-guarded AU objective. Third, because every
added module can overfit validation, the ablation path must remain explicit. We
make the following contributions.

\begin{itemize}[leftmargin=1.3em,itemsep=2pt,topsep=2pt]
  \item \textbf{\textit{A data-efficient pipeline.}} The pipeline built from
  frozen AffectNet-supervised face encoders, light task heads, and ablation-guided
  component selection reaches a validation $\mathcal{P}$ of 1.7302 without
  training a new large-scale face foundation model (\cref{sec:method}).
  \item \textbf{\textit{The encoder is not always the bottleneck.}} We show that
  supervised frozen features are strong for expression and valence/arousal, and
  give a recipe that recovers the after-encoder gains:
  cross-task feature fusion (\cref{sec:fusion}), a LoRA-adapted action-unit
  encoder trained under a collapse-guarded loss (\cref{sec:ft}), and per-unit
  expert routing that folds in external DISFA data without the domain collapse
  of sequential fine-tuning (\cref{sec:disfa}).
  \item \textbf{\textit{A compact ablation study.}} We report the useful
  validation gains and the negative results that mark the edge of what frozen
  transfer can do on this data (\cref{sec:results}).
\end{itemize}

%% file: sections/02-related-work.tex
\section{Related Work}
\label{sec:related}
\paragraph{\normalfont\bfseries\itshape ABAW multi-task learning.}
The challenge has run valence/arousal estimation, expression recognition, and
action-unit detection, singly and jointly, across successive
editions~\cite{kollias2020analysing,kollias2021analysing,kollias2022abaw,kollias2023abaw2,kollias2023abaw,kollias20246th,kollias20247th,kollias2025advancements,kollias2025emotions,kollias2026affect},
all on the partially labelled \mbox{s-Aff-Wild2} subset of
Aff-Wild2~\cite{kollias2019expression,kollias2019deep,kollias2021affect,kollias2019face},
where many frames carry only a subset of the three label types; here, the prefix
``s'' denotes the static selected-frame version used for MTL. The organizers
also study label-distribution matching for this heterogeneous
structure~\cite{kollias2024distribution,kollias2021distribution} and release
in-the-wild analysis toolkits~\cite{kollias2024behaviour4all}. A 7th-edition
MTL system~\cite{liu2024progressive} combined a masked-autoencoder extractor,
cross-task feature fusion, an LSTM temporal module, and a progressive
separate-then-joint training schedule; its ablation shows action-unit features
lifting expression and valence/arousal, most on arousal, while fusion slightly
hurts the action-unit branch itself. These observations motivate our own fusion
ablations.
HSEmotion~\cite{savchenko2024hsemotion,savchenko2022video} instead used small
AffectNet backbones with per-unit adaptive thresholds and inference-time
temporal smoothing, the last of which we also adopt.

\paragraph{\normalfont\bfseries\itshape Frozen backbones, external data, and rare classes.}
DDAMFN~\cite{zhang2023ddamfn} and AffectNet-supervised EfficientNet
models~\cite{mollahosseini2017affectnet,tan2019efficientnet} give strong
in-domain expression and valence/arousal features that we use without further
training; a masked autoencoder~\cite{he2022masked,zhang2023multimodal} instead
serves here as a fine-tuning target for action units, not a frozen feature
source. External corpora are the usual lever for rare action units: BP4D is a
common choice in prior work, and we use
DISFA~\cite{mavadati2013disfa,mavadati2012automatic} for the positives
Aff-Wild2 lacks, though we find lab-recorded data transfers poorly when used to
initialise the in-the-wild encoder and route it in per column instead.
Because rare expressions and sparse AU positives affect the macro-$F_1$ terms
directly, we report class-wise and unit-wise validation diagnostics to expose
which labels drive the remaining errors.

%% file: sections/03-method.tex
\section{Method}
\label{sec:method}

\begin{figure*}[t]
  \centering
  \includegraphics[width=\textwidth]{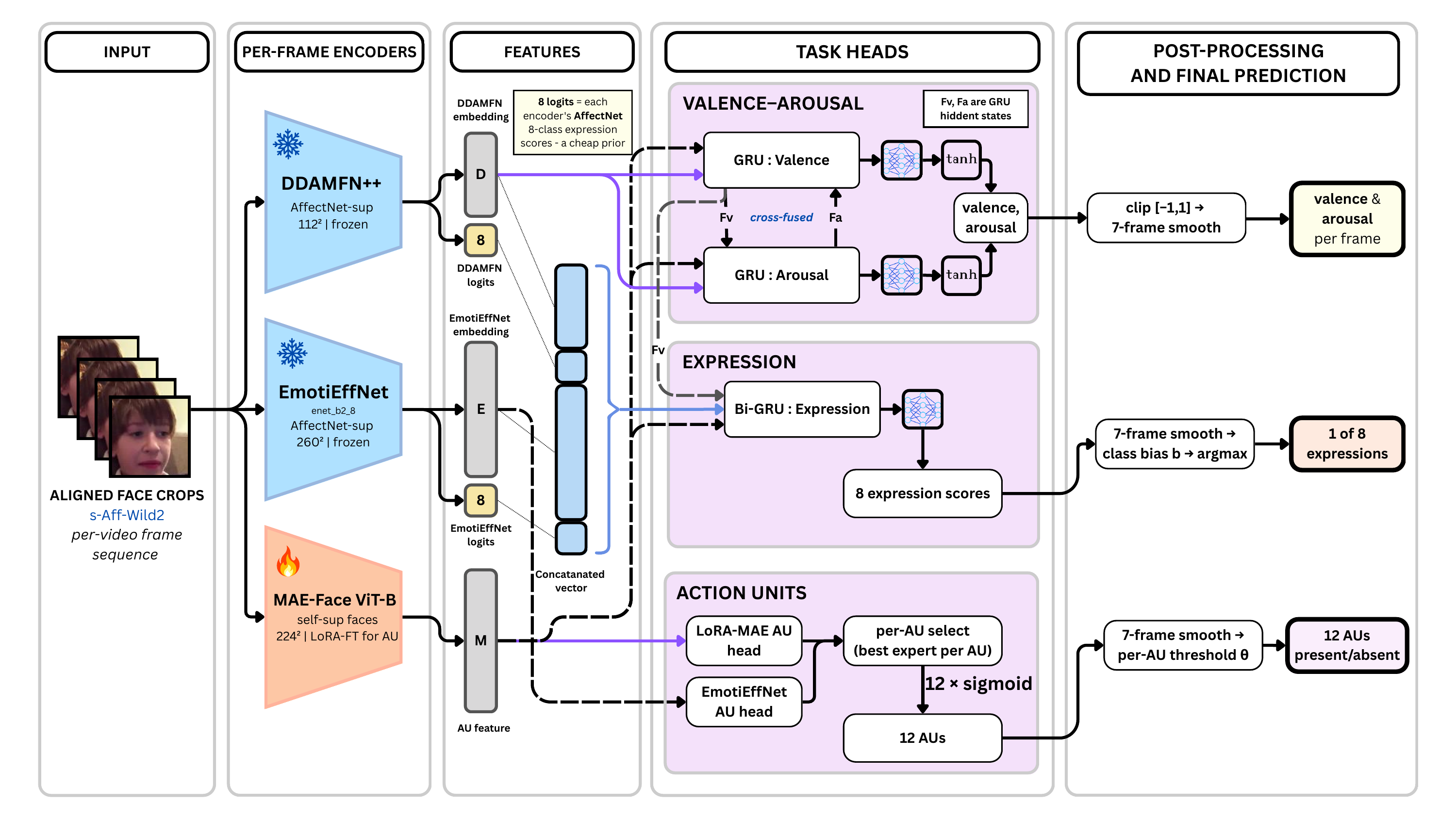}
  \caption{Overview of AffectFuse. Frozen AffectNet-supervised encoders and an
  AU-adapted MAE-Face encoder produce per-frame features. The AU representation
  is fused into the temporal VA and EXPR heads, while the AU branch remains
  independent. A seven-frame moving average and task-specific calibration
  produce the final predictions.}
  \label{fig:system-overview}
\end{figure*}

\subsection{Candidate Pipeline}

The system overview is shown in \cref{fig:system-overview}. It is a staged
collection of task-specific candidate models rather than a single end-to-end
multi-task network. Pretrained encoders first produce per-frame features;
lightweight heads, temporal models, fusion modules, and ensembles are then
evaluated for each task. For validation metric $M_t$, the selected predictor for
task $t$ is
\begin{equation}
  f_t^\star=\arg\max_{f\in\mathcal{C}_t} M_t(f),
  \qquad
  \mathcal{P}=\sum_t M_t(f_t^\star),
  \label{eq:candidate-selection}
\end{equation}
where $M_{\mathrm{VA}}$ is mean valence/arousal CCC~\cite{lin1989concordance},
$M_{\mathrm{EXPR}}$ is eight-class macro-$F_1$, and $M_{\mathrm{AU}}$ is
mean $F_1$ over 12 action units. This task-wise selection is used to organize
the ablation study and to choose the final validation configuration.

\subsection{Frozen Affective Experts and Features}
\label{sec:backbones}

We use three complementary pretrained face encoders
(\cref{tab:backbones}). The first is \texttt{enet\_b2\_8}, an
EfficientNet-B2 from the EmotiEffNet family (\textsc{Enet})
\cite{tan2019efficientnet,savchenko2022video,savchenko2023mtemotieffnet}; at $260\times260$ it returns an
embedding $e_i\in\mathbb{R}^{1408}$ and eight expression logits
$p_i^E$. DDAMFN~\cite{zhang2023ddamfn} operates at $112\times112$ and returns
$d_i\in\mathbb{R}^{512}$ and logits $p_i^D\in\mathbb{R}^{8}$. Both models are
AffectNet-supervised and remain frozen. Because their output classes use
different conventions, both logit vectors are permuted to the Aff-Wild2
challenge order before use.

The main expression representation is
\begin{equation}
  x_i^e=
  [\,d_i\,\|\,\widetilde{p}_i^D\,\|\,e_i\,\|\,\widetilde{p}_i^E\,]
  \in\mathbb{R}^{1936},
  \label{eq:xe}
\end{equation}
where the tildes denote remapped logits. Initial VA, EXPR, and AU heads use $d_i$,
$x_i^e$, and $e_i$, respectively. The third encoder is a self-supervised
MAE-Face ViT-B~\cite{he2022masked,zhang2023multimodal,dosovitskiy2021image}. At $224\times224$, its
class token and mean patch token are concatenated into a 1536-dimensional
feature. MAE-Face is weak as a frozen affect candidate but becomes the AU
expert after the adaptation in \cref{sec:ft}. Frozen outputs are cached once;
adapted MAE features are cached after each fine-tuning stage.

\begin{table}[t]
  \centering
  \caption{Pretrained encoders and per-frame outputs.}
  \label{tab:backbones}
  \setlength{\tabcolsep}{3pt}
  \small
  \begin{tabular}{@{}lccrr@{}}
    \toprule
    Encoder & Pretraining & Input & Emb. & EXPR logits \\
    \midrule
    \textsc{Enet}-B2 & AffectNet & $260^2$ & 1408 & 8 \\
    DDAMFN & AffectNet & $112^2$ & 512 & 8 \\
    MAE-Face ViT-B & self-sup. faces & $224^2$ & 1536 & -- \\
    \bottomrule
  \end{tabular}
\end{table}

\subsection{Initial Heads and Objectives}
\label{sec:heads}

The initial candidates are shallow MLP heads trained on cached features. VA
uses a two-output, tanh-bounded regressor on $d_i$ with MSE plus a CCC loss for
each axis. EXPR uses an eight-way softmax head on $x_i^e$, optimized with
class-weighted cross-entropy and label smoothing. AU uses 12 independent
sigmoid outputs on $e_i$ and a masked asymmetric loss
(ASL~\cite{benbaruch2021asymmetric}), a multi-label cross-entropy that
down-weights easy negatives to counter the heavy positive/negative imbalance of
rare units.

Missing annotations contribute no loss or metric: VA frames marked $-5$ and
EXPR labels marked $-1$ are ignored, while AU labels marked $-1$ are masked
independently for each column. Thus a frame can supervise any available task
or AU without requiring a complete multi-task annotation.

\subsection{Temporal Candidate Heads}
\label{sec:temporal}

For temporal candidates, frames are grouped by video, sorted by frame index,
and partitioned into fixed-length windows; the last short window is padded by
repeating its final frame. Gated recurrent unit (GRU)~\cite{cho2014learning} and bidirectional GRU heads are evaluated for VA
and EXPR. Bidirectional context is valid because inference is offline and the
complete recording is available. We also evaluate the Temporal Convergence
Module (TCM) described in~\cite{liu2024progressive}, a two-layer long short-term
memory (LSTM)~\cite{hochreiter1997long} followed
by LeakyReLU, on the donor-augmented VA and EXPR features.

Temporal modeling did not provide the same benefit for AU, so retained
sequence candidates focus on VA and EXPR. Window length is selected as a
task-level hyperparameter: increasing $L$ from 32 to 48 substantially improved
the fused temporal candidates, and the final implementation uses $L=48$ for
both tasks.

\subsection{Parameter-Efficient AU Adaptation}
\label{sec:ft}

Frozen features gave limited gains for AU, so we adapt MAE-Face on Aff-Wild2
AU labels. LoRA updates are inserted into the attention QKV/output projections
and the two MLP projections of every transformer block,
\begin{equation}
  W'=W+\frac{\alpha}{r}BA,
  \qquad r=16,\quad \alpha=32.
  \label{eq:lora}
\end{equation}
Here $r$ is the adapter rank, which fixes the size of $B\in\mathbb{R}^{d\times r}$
and $A\in\mathbb{R}^{r\times k}$ and hence the parameter budget, while $\alpha/r$
sets the magnitude of the update; we keep LoRA's two-constant
form~\cite{hu2022lora} so that $r$ can be retuned without rescaling the update.
All LayerNorm parameters, the AU head, and the complete final four transformer
blocks are also optimized. This leaves $31.53$M of $89.0$M parameters
trainable.

Optimization uses layer-wise learning-rate decay of $0.65$, linear warmup
followed by cosine decay, gradient clipping, image augmentation, and an
exponential moving average with decay $0.999$. The masked AU loss is
\begin{equation}
  \mathcal{L}_{\mathrm{AU}}
  =\mathcal{L}_{\mathrm{ASL}}
  +\lambda(\tau)\mathcal{L}_{\mathrm{sF1}},
  \label{eq:au-loss}
\end{equation}
where $\mathcal{L}_{\mathrm{sF1}}$ is a bounded sigmoid-$F_1$ surrogate~\cite{benedict2021sigmoidf1} with
$\beta=2$, and $\lambda(\tau)$ is ramped in after warmup. An earlier unbounded
surrogate collapsed to an all-positive solution. During the ramp, the metric
term is therefore disabled for the remainder of training if the validation
predicted-positive rate exceeds $0.80$ or AU $F_1$ falls below $0.42$.

\subsection{Cross-Task Feature Fusion}
\label{sec:fusion}

Let $F_i^{au}\in\mathbb{R}^{1536}$ denote the feature from an AU-adapted
MAE-Face encoder. AU serves as a donor to the other tasks. For EXPR, a BiGRU
first consumes $[x_i^e\|F_i^{au}]$; a second candidate also receives the
pre-classifier valence representation $F_i^v$:
\begin{equation}
  z_i^{e,1}=[\,x_i^e\|F_i^{au}\,],
  \qquad
  z_i^{e,2}=[\,x_i^e\|F_i^{au}\|F_i^v\,].
  \label{eq:expr-fusion}
\end{equation}

VA is split into separate valence and arousal heads. Both first use
$[d_i\|F_i^{au}]$; their pre-classifier states are then cross-fused:
\begin{equation}
  z_i^v=[\,d_i\|F_i^{au}\|F_i^a\,],
  \qquad
  z_i^a=[\,d_i\|F_i^{au}\|F_i^v\,].
  \label{eq:va-fusion}
\end{equation}
The LSTM-based convergence module is added as another VA/EXPR candidate on the
donor-augmented features. We do not feed VA or EXPR features back into the AU
branch, since AU fusion reduced validation performance in the corresponding
ablations.

\subsection{External AU Transfer and Per-Unit Harvest}
\label{sec:disfa}

We prepare DISFA as 75,087 landmark-aligned $224\times224$ crops from 27
subjects. It annotates 8 of the 12 challenge AUs; AU7, AU10, AU23, and AU24 are
stored as $-1$ and masked. In Stage A, MAE-Face is LoRA-adapted on DISFA using
a subject-disjoint holdout. Stage B continues that model on all Aff-Wild2 AUs.
The resulting direct AU candidates do not surpass the Aff-Wild2-only expert,
consistent with a laboratory-to-in-the-wild domain gap, but they remain in the
AU pool and are reused as VA/EXPR donors.

We also train the feature-level candidate \texttt{ftmix} on
$[x_i^e\|F_i^{\mathrm{AW}}\|F_i^{\mathrm{D}\rightarrow\mathrm{AW}}]$, where
the two MAE features come from the Aff-Wild2-only and DISFA-initialized Stage-B
encoders. Although \texttt{ftmix} is not the best complete AU predictor, it
remains available to the per-unit harvest. The fusion stack is likewise rerun
with the DISFA-initialized donor and with both MAE donors concatenated.

Finally, \texttt{aumix} selects an expert independently for each AU column:
\begin{equation}
  r_j^\star=\arg\max_{r\in\mathcal{R}}F_{1,j}^{\mathrm{val}}(r),
  \qquad
  \hat{y}_{i,j}^{\mathrm{AU}}=f_{r_j^\star,j}(x_i).
  \label{eq:aumix}
\end{equation}
Each route retains its expert's fitted threshold. This design follows the
official AU metric, which averages column-wise $F_1$, and lets different AU
labels use different sources when their validation behavior differs.

\subsection{Variance-Reduction Ensembles}
\label{sec:phase1}

The final stage expands the pools with variance-reduced predictors. We rerun
the fusion stack with three random seeds and average predictions from matching
candidate families. We then form top-$K$ cross-candidate ensembles by
averaging the $K=3$ highest-scoring validation candidates for each task.
Horizontal flip test-time augmentation, defined by averaging original and
flipped predictions, is evaluated in the same way.

All variants re-enter the per-task argmax in \cref{eq:candidate-selection}.
The final pool selects a seed ensemble for VA and a top-three ensemble for
EXPR, while AU retains \texttt{aumix}. Flip augmentation is not used because it
lowers validation scores for all three tasks.

\subsection{Post-Processing}
\label{sec:postproc}

Let $S_7$ denote a centered seven-frame moving average applied separately
within each video. The selected outputs are converted to predictions as
\begin{equation}
  \begin{aligned}
    \widehat{\mathbf{y}}_i^{\mathrm{VA}}
      &=\operatorname{clip}\!\left(S_7(\mathbf{y}^{\mathrm{VA}})_i,-1,1\right),\\
    \widehat{c}_i
      &=\arg\max_c\left\{S_7(\ell^{\mathrm{EXPR}})_{i,c}+b_c\right\},\\
    \widehat{a}_{i,j}
      &=\mathbf{1}\!\left[S_7\!\left(\sigma(\ell^{\mathrm{AU}}_{i,j})\right)
        \geq\theta_j\right].
  \end{aligned}
  \label{eq:postproc}
\end{equation}
The EXPR bias $b\in\mathbb{R}^{8}$ and the 12 AU thresholds $\theta_j$ are
fitted on validation after smoothing. These parameters, together with the
selected candidate and AU route map, are saved and reused unchanged at
inference.

%% file: sections/04-experimental-setup.tex
\section{Experiments}
\label{sec:experiments}

\subsection{Dataset}
\label{sec:dataset}

For the MTL track, a total of 142{,}382 training images and 26{,}876 validation
images from \mbox{s-Aff-Wild2}, the static selected-frame version of
Aff-Wild2~\cite{kollias2018affwild2}, are provided by the organizers. After 
removing duplicate entries from the annotation files, the training and validation
datasets consist of 141,431 and 26,666 images, respectively. Rows with
invalid labels are masked in the corresponding task
losses using the official sentinels.

\paragraph{Extra Data.}
We preprocessed \textbf{DISFA}~\cite{mavadati2013disfa} into 75{,}087
landmark-aligned crops over 27 subjects; 8 of our 12 AUs are coded there. We
additionally acquired and audited \textbf{AffectNet+}~\cite{pourramezanfard2025affectnetplus}.

\subsection{Implementation Details}

We used a multi-backbone, multi-resolution feature extraction pipeline for the MTL track.
EfficientNet-B2 was evaluated at $260\times260$, DDAMFN at $112\times112$, and MAE-Face at $224\times224$, matching each backbone's native input setting. The prediction heads were optimized with AdamW using task-appropriate objectives: CCC loss for VA, cross-entropy for EXPR, and asymmetric binary loss for AU detection. All experiments were conducted on an NVIDIA RTX Pro 6000 GPU; complete hyperparameter settings, including batch size, learning rate, scheduler, and optimizer configuration, are provided in the released code.

\subsection{Experimental Results}
\label{sec:results}

\begin{figure}[H]
\centering
\includegraphics[width=0.75\linewidth]{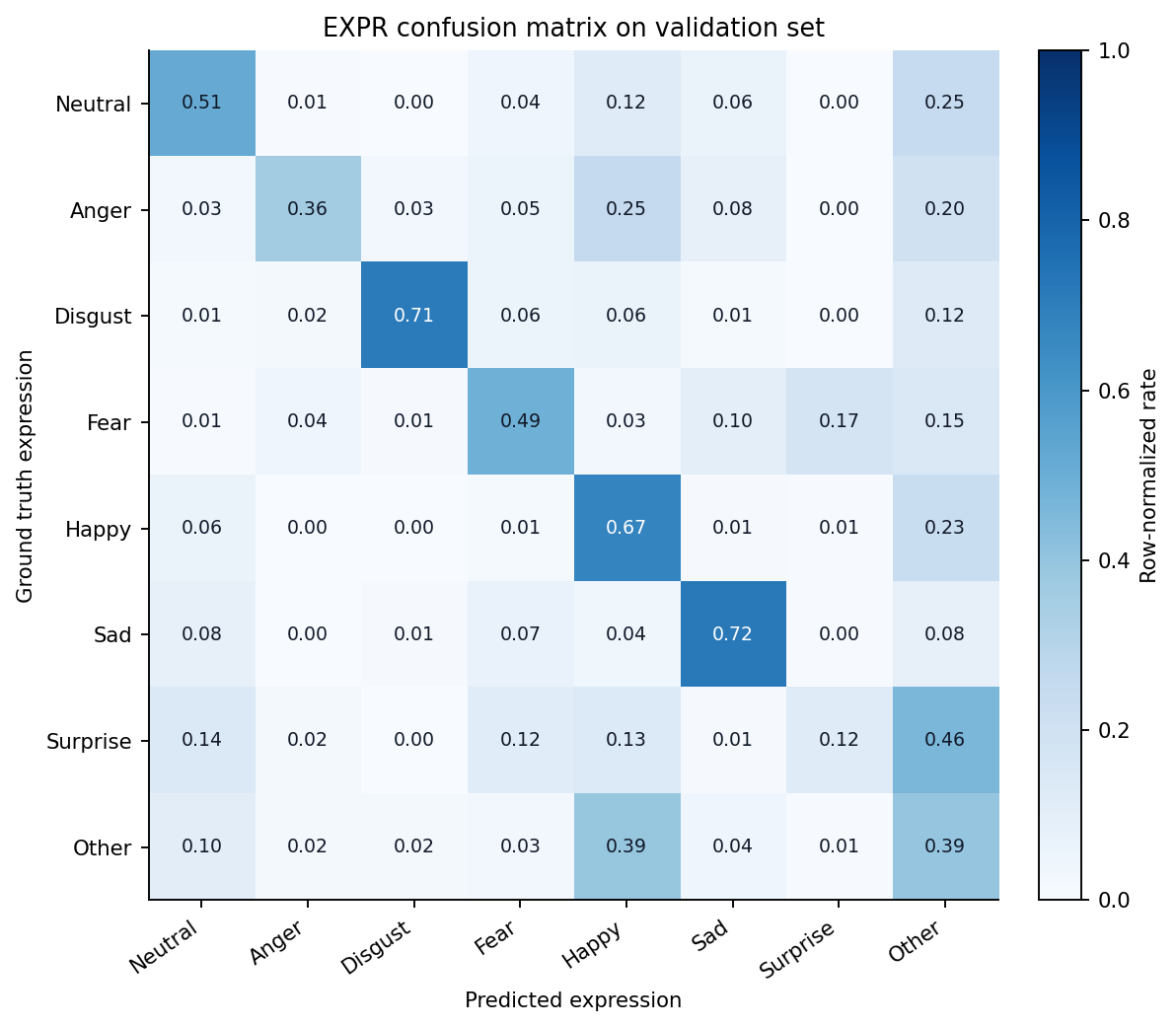}
\caption{EXPR confusion matrix on validation set.}
\label{fig:expr_confusion}
\end{figure}

\paragraph{Analysis on the Validation Dataset.}
We evaluate our final approach to the MTL challenge on the official validation
dataset. Table~\ref{tab:main} gives the main validation score. Figure~\ref{fig:expr_confusion}
shows the EXPR confusion matrix on the same split.

\begin{table}[H]
\centering
\caption{Quantitative results on the officially provided validation dataset.}
\label{tab:main}
\setlength{\tabcolsep}{8pt}
\footnotesize
\begin{tabular}{lcccc}
\toprule
Method & Avg. CCC & $F_{\mathrm{expr}}$ & $F_{\mathrm{AUs}}$ & $P$ \\
\midrule
Organizer score & -- & -- & -- & 0.4500 \\
\textbf{Ours} & \textbf{0.6596} & \textbf{0.5097} & 0.5609 & \textbf{1.7302} \\
\bottomrule
\end{tabular}
\end{table}

Table~\ref{tab:progression} summarizes the complete validation path. Starting
from frozen backbone features and plain task heads, the ablation adds one family
of candidates at each stage and reports the selected task-wise configuration.
The largest gains come
from temporal modeling, cross-task feature fusion, and the final variance
reduction step, while AU-specific changes mainly stabilize the action-unit
branch.
\begin{table}[H]
\centering
\caption{Validation progression.}
\label{tab:progression}
\setlength{\tabcolsep}{3pt}\footnotesize

\begin{tabular}{lccccr}
\toprule
Stage & VA & EXPR & AU & $P$ & $\Delta$ \\
\midrule
Frozen backbones, plain heads            & 0.5294 & 0.3948 & 0.5287 & 1.4529 & -- \\
AU on concat. features + ASL             & 0.5870 & 0.4165 & 0.5401 & 1.5436 & +0.091 \\
AU: LoRA fine-tuned MAE                  & 0.5870 & 0.4165 & 0.5573 & 1.5608 & +0.017 \\
GRU tuning                               & 0.6188 & 0.4314 & 0.5573 & 1.5969 & +0.036 \\
Decoupled window + BiGRU                 & 0.6235 & 0.4477 & 0.5573 & 1.6284 & +0.032 \\
\textbf{Cross-task feature fusion}       & 0.6260 & 0.4841 & 0.5573 & \textbf{1.6674} & \textbf{+0.039} \\
DISFA + re-fusion                        & 0.6369 & 0.4841 & 0.5573 & 1.6783 & +0.011 \\
Per-AU mixture of experts                & 0.6361 & 0.4838 & 0.5659 & 1.6858 & +0.008 \\
\textbf{Temporal window $32\!\to\!48$}   & 0.6526 & 0.5035 & 0.5609 & \textbf{1.7170} & \textbf{+0.031} \\
Multi-seed + top-3 ensembling            & 0.6596 & 0.5097 & 0.5609 & \textbf{1.7302} & +0.013 \\
\bottomrule
\end{tabular}
\end{table}

The following ablations unpack the main rows of Table~\ref{tab:progression}:
backbone choice, MAE-Face adaptation, fusion, external AU data, temporal window
length, and ensembling.

\paragraph{Impact of Backbone Diversity.}
Table~\ref{tab:backbone} compares the frozen feature sets used before temporal
modeling. The two AffectNet backbones gave the best pair; extra frozen backbones
did not improve validation scores.

\begin{table}[!ht]
\centering
\caption{Backbone comparison.}
\label{tab:backbone}
\small
\resizebox{\columnwidth}{!}{%
\begin{tabular}{llccl}
\toprule
Candidate  & EXPR & AU & Decision \\
\midrule
\textbf{EfficientNet-B2+DDAMFN++}  & \textbf{0.4165} & \textbf{0.5401} & used \\
+ POSTER++~\cite{mao2023poster}  & 0.4164 & 0.5373 & no gain \\
MAE-Face, frozen  & 0.3770 & -- & low EXPR \\
\bottomrule
\end{tabular}%
}
\end{table}

POSTER++ changed EXPR by only $-0.0001$ and reduced AU, so it was not used in
the final frozen feature set. Frozen MAE-Face was also weaker for EXPR before
task-specific adaptation.

\paragraph{Impact of Fine-tuning the MAE Feature Extractor.}
Table~\ref{tab:ft} shows the AU result after LoRA fine-tuning MAE-Face.

\begin{table}[!ht]
\centering
\caption{LoRA fine-tuning MAE-Face for AU.}
\label{tab:ft}
\setlength{\tabcolsep}{8pt}\small
\begin{tabular}{ccccc}
\toprule
Unfreeze & Params & AU & $P$ & Gain \\
\midrule
2 & 15.9\,M & 0.5485 & 1.5520 & +0.0084 \\
4 & 30.1\,M & 0.5519 & 1.5554 & +0.0118 \\
\textbf{6} & ${\sim}37$\,M & \textbf{0.5573} & \textbf{1.5608} & \textbf{+0.0172} \\
8 & ${\sim}44$\,M & 0.5544 & 1.5579 & +0.0143 \\
\bottomrule
\end{tabular}
\end{table}

\begin{figure}[!ht]
\centering
\includegraphics[width=0.88\linewidth]{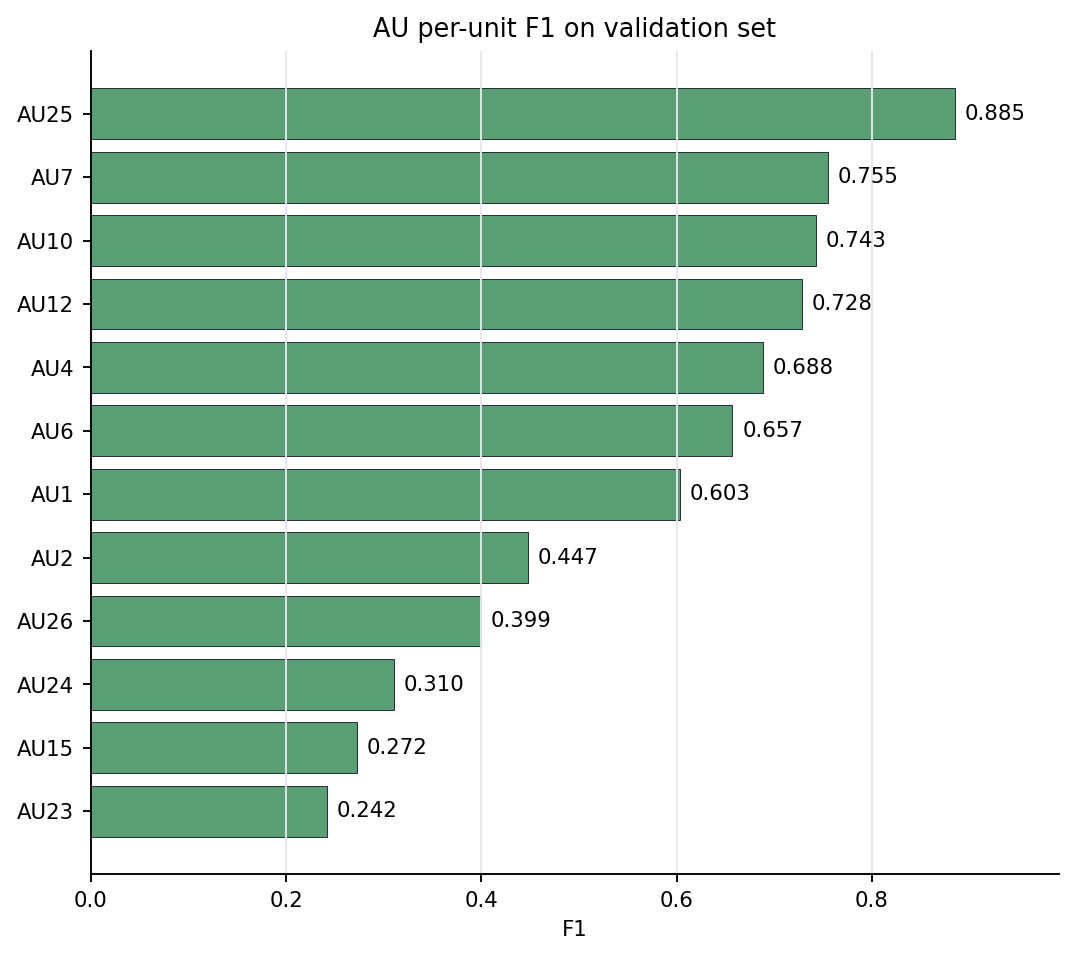}
\caption{AU per-unit F1 on validation set.}
\label{fig:au_f1}
\end{figure}

Six unfrozen blocks gave the best AU score. The same MAE fine-tuning did not help
EXPR or VA (Table~\ref{tab:ftall}); the EXPR run saturated at $0.4012$, below our
frozen feature ensemble. Figure~\ref{fig:au_f1} shows the AU
stats at the unit level on the validation set.

\begin{table}[!ht]
\centering
\caption{Fine-tuning MAE-Face for EXPR and VA.}
\label{tab:ftall}
\setlength{\tabcolsep}{3pt}\small
\begin{tabular}{lcccc}
\toprule
Task & ft-MAE & + concat & + temp. & \textbf{Frozen best} \\
\midrule
EXPR & 0.3667 & 0.3909 & \textbf{0.4012} & \textbf{0.4477} \\
VA & 0.4332 & 0.4429 & 0.5579 & \textbf{0.6235} \\
\bottomrule
\end{tabular}
\end{table}

\paragraph{Impact of Feature Fusion.}
Table~\ref{tab:fusion} shows the useful candidates introduced at the fusion
stage.

\begin{table}[!ht]
\centering
\caption{Fusion-stage candidates.}
\label{tab:fusion}
\begin{tabular}{lccc}
\toprule
Task & Before & Best candidate & Gain \\
\midrule
EXPR & 0.4477 & \textbf{0.4841} ($F_{au}+F_v$) & \textbf{+0.0364} \\
VA & 0.6235 & \textbf{0.6260} (TCM) & +0.0025 \\
\bottomrule
\end{tabular}
\end{table}

Fusion mainly helped EXPR. AU features alone were not enough, but combining AU
and valence features raised EXPR from $0.4477$ to $0.4841$. VA improved slightly
through the temporal convergence module rather than cross-task donors. AU was
kept unfused because all fusion variants were below $0.5573$.

\paragraph{Impact of External AU Data.}
Table~\ref{tab:disfa} evaluates sequential DISFA-to-Aff-Wild2 AU transfer,
where the final comparison is measured on the official Aff-Wild2 validation
split.

\begin{table}[!ht]
\centering
\caption{Impact of DISFA sequential transfer.}
\label{tab:disfa}
\begin{tabular}{lc}
\toprule
Training path & AU \\
\midrule
Stage-A, held-out DISFA subjects & \textbf{0.7280} \\
Stage-B on Aff-Wild2 (lr $\times0.4$) & 0.5451 \\
\textbf{Aff-Wild2-only training} & \textbf{0.5573} \\
\bottomrule
\end{tabular}
\end{table}

DISFA pre-training worked on held-out DISFA subjects, but after transfer to
Aff-Wild2 it did not surpass the Aff-Wild2-only AU model. We therefore did not
use this branch as the final AU classifier; its useful role was as an additional
donor in the later re-fusion step.

\paragraph{Impact of the Temporal Module.}
Table~\ref{tab:gru} shows the effect of window size and hidden dimension before
fusion.

\begin{table}[!ht]
\centering
\caption{Temporal module before fusion.}
\label{tab:gru}
\begin{tabular}{ccc}
\toprule
$W$ / $H$ & VA & EXPR \\
\midrule
32 / 256 & 0.6016 & 0.4149 \\
\textbf{32 / 512} & 0.6188 & \textbf{0.4314} \\
32 / 768 & 0.6086 & 0.4295 \\
\textbf{48 / 512} & \textbf{0.6235} & 0.4105 \\
\bottomrule
\end{tabular}
\end{table}

Hidden size $512$ was best in this sweep. A longer window helped VA but reduced
raw EXPR, so later heads were allowed to use task-specific windows.

\begin{table}[!ht]
\centering
\caption{Temporal window after fusion.}
\label{tab:win48}
\begin{tabular}{cccr}
\toprule
Task & $W{=}32$ & $\mathbf{W{=}48}$ & $\Delta$ \\
\midrule
EXPR & 0.4838 & \textbf{0.5035} & +0.020 \\
VA & 0.6361 & \textbf{0.6526} & +0.017 \\
AU & 0.5659 & 0.5609 & $-0.005$ \\
\textbf{$P$} & 1.6858 & \textbf{1.7170} & \textbf{+0.031} \\
\bottomrule
\end{tabular}
\end{table}

On the fused pipeline, increasing the window to $48$ improved VA and EXPR and
raised $P$ by $+0.031$. This shows that the best temporal length depends on the
feature representation, not only on the task.

\paragraph{Impact of Variance Reduction.}
Table~\ref{tab:phase1} reports the final ensembling step.

\begin{table}[!ht]
\centering
\caption{Multi-seed, top-$K$, and flip-TTA.}
\label{tab:phase1}
\setlength{\tabcolsep}{3pt}\small
\begin{tabular}{lcccl}
\toprule
Task & Before & Seed/top-$K$ & Flip-TTA & Selected \\
\midrule
VA & 0.6526 & \textbf{0.6596} & 0.6577 & 3-seed ens. \\
EXPR & 0.5034 & \textbf{0.5097} & 0.5040 & top-3 ens. \\
AU & 0.5609 & 0.5609 & 0.5606 & unchanged \\
\textbf{$P$} & 1.7168 & \textbf{1.7302} & -- & \\
\bottomrule
\end{tabular}
\end{table}

Multi-seed and top-$K$ ensembling improved VA and EXPR. Flip-TTA was lower for
all tasks, likely because expressions and AUs can be asymmetric after a
horizontal flip. We therefore excluded flip-TTA and retained the multi-seed/top-$K$
configuration, which gave the best validation score of $P=1.7302$.

%% file: sections/06-conclusion.tex
\section{Conclusion}
\label{sec:conclusion}
A data-efficient pipeline of frozen AffectNet-supervised face encoders,
cross-task feature fusion, a LoRA-adapted action-unit expert, and per-unit
routing of external data reaches a validation $\mathcal{P}$ of 1.7302 on the
11th ABAW MTL track, with strong valence/arousal and expression performance,
without training a new large-scale face foundation model or relying on a
licensed AU corpus. The system is built through a staged ablation framework that
compares backbone, fusion, temporal, AU-adaptation, and ensembling choices under
the same validation protocol. Two findings carry beyond our entry. The encoder is not always the bottleneck for
valence/arousal and expression, since a fine-tuned autoencoder saturates below
the frozen supervised ensemble; effort is better spent after the encoder, on
action units. The remaining validation errors point
to action-unit coverage and expression-class confusion as the main residual
issues. Strong frozen features combined thoughtfully after the encoder, and
selected through explicit ablations, offer a reproducible route without training
a new large-scale face foundation model.